\documentclass[a4paper, 10pt, conference]{ieeeconf}      
\usepackage{FG2024}

\FGfinalcopy 

\IEEEoverridecommandlockouts                              
\let\labelindent\relax
\usepackage{enumitem}
\usepackage{todonotes}

\usepackage{subcaption}
\usepackage{float}
\usepackage{dblfloatfix} 
\usepackage{makecell}

\usepackage[utf8]{inputenc}

\usepackage{mathtools}

\usepackage{amsmath}
\usepackage{amsfonts}
\newcommand{\matr}[1]{\mathbf{#1}}
\newcommand{\vect}[1]{\textbf{#1}}

\DeclareMathOperator*{\argmax}{arg\,max}

\newcommand\Alpha{\mathrm{A}}

\def\FGPaperID{251} 

\title{\LARGE \bf
Boosting Gesture Recognition with an Automatic Gesture
Annotation Framework
}

\author{\parbox{16cm}{\centering
    {\large Junxiao Shen$^{1,2}$, Xuhai Xu$^{1,3}$, Ran Tan$^1$, Amy Karlson$^1$, and Evan Strasnick$^1$}\\
    {\normalsize
    $^3$ Department of Electrical Engineering and Computer Science, Massachusetts Institute of Technology\\
    $^2$ Department of Engineering, University of Cambridge\\
    $^1$ Reality Labs Research, Meta} }
}

\usepackage{fancyhdr}
\begin{document}

\ifFGfinal
\thispagestyle{empty}
\pagestyle{empty}
\else
\author{Anonymous FG2024 submission\\ Paper ID \FGPaperID \\}
\pagestyle{plain}
\fi
\maketitle

\thispagestyle{fancy}

\begin{abstract}
Training a real-time gesture recognition model heavily relies on annotated data. However, manual data annotation is costly and demands substantial human effort. In order to address this challenge, we propose a framework that can automatically annotate gesture classes and identify their temporal ranges. Our framework consists of two key components: (1) a novel annotation model that leverages the Connectionist Temporal Classification (CTC) loss, and (2) a semi-supervised learning pipeline that enables the model to improve its performance by training on its own predictions, known as pseudo labels. These high-quality pseudo labels can also be used to enhance the accuracy of other downstream gesture recognition models.
To evaluate our framework, we conducted experiments using two publicly available gesture datasets. Our ablation study demonstrates that our annotation model design surpasses the baseline in terms of both gesture classification accuracy (3-4\% improvement) and localization accuracy (71-75\% improvement). Additionally, we illustrate that the pseudo-labeled dataset produced from the proposed framework significantly boosts the accuracy of a pre-trained downstream gesture recognition model by 11-18\%.
We believe that this annotation framework has immense potential to improve the training of downstream gesture recognition models using unlabeled datasets.

\end{abstract}

\section{INTRODUCTION}

\begin{figure*}
\centering
  \includegraphics[width=0.75\linewidth]{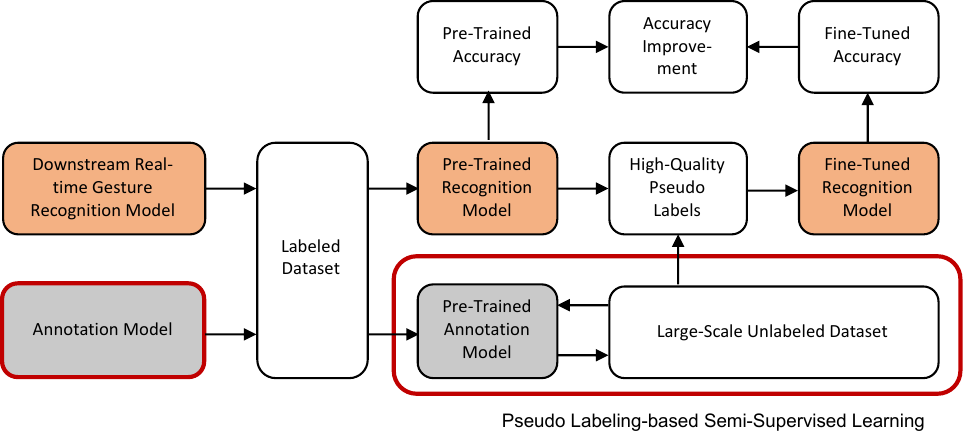}
  \caption{Our framework provides an automatic gesture annotation solution. The red borders highlight our main contribution. The framework consists of two components: (1) a novel annotation model that utilizes Connectionist Temporal Classification (CTC) loss (see Figure~\ref{fig:annotation_model}), and (2) a semi-supervised pipeline that improves the model's performance by training on its own predictions, i.e., pseudo labels.
  In real-life (open-world), the gesture annotation framework operates as follows: first, a real-time gesture recognition model and our proposed annotation model are pre-trained on a small labeled dataset. Next, the annotation model is integrated into the pseudo-labeling process, where it produces pseudo labels by annotating an unlabeled dataset, which is then used to augment the annotation model. After pseudo-labeling is complete, the final high-quality pseudo-labels are used to fine-tune the pre-trained real-time gesture recognition model.}
  \label{fig:illustration}
\end{figure*}



Hand gesture recognition is a widely applied technology in various fields such as augmented reality (AR)~\cite{reifinger2007static}, and virtual reality (VR)~\cite{sagayam2017hand,Shen2022Gesture}.
Various modalities of hand gesture data can be utilized for recognition, such as RGB, optical flow, depth, IR, IR-disparity, and 2D/3D skeletons. Contemporary AR and VR head-mounted devices (HMDs) track the user's hands, providing access to real-time 3D hand skeleton data. In this paper, our focus is on using 3D hand skeleton data. This choice is motivated by the widespread availability of 3D hand skeleton data in modern HMDs and its compatibility across different sensors and devices~\cite{Shen2022Gesture}.

A typical pipeline to build a robust gesture recognition model usually starts with data collection, followed by data annotation and model training.
The annotation step involves two steps: (1) marking the class of the gesture, and (2) localizing the time range of a gesture.
State-of-the-art gesture recognition methods are primarily based on deep-learning techniques~\cite{liu2020decoupled,jaramillo2020real}, which require a significant amount of labeled data for training~\cite{najafabadi2015deep}. 
Insufficient training data can result in overfitting or a failure to learn for a deep neural-network model ~\cite{shen2021imaginative}.
Unfortunately, annotating data can be both time-consuming and costly, especially for open-world gesture recognition~\cite{Shen2024TowardsOG}.
Additionally, many state-of-the-art gesture recognition methods require strongly-segmented data (frame-wise segmentation, i.e., Step 2 of annotation) for high-quality training~\cite{jaramillo2020real,chen2019construct,liu2020decoupled}, which needs more intense human manual efforts and adds another layer of expense to the data annotation process.

Previous research has focused on addressing the issue of data sparsity in gesture recognition. Various techniques for data augmentation have been proposed, such as the use of a Generative Adversarial Network (GAN) for realistic transformations on gesture skeleton data~\cite{shen2021imaginative,Shen2021SimulatingRH}, and the use of unrealistic distortions like cutout~\cite{devries2017improved} and mixup~\cite{zhang2017mixup} to regularize neural network training. Another recent work by Xu et al.~\cite{Xu2022Enabling} leveraged a few-shot learning framework to reduce the dataset size requirement.
However, while these techniques help alleviate the issue of data sparsity, they all rely on an existing annotated dataset, which still suffers from the high cost of data annotation. There is very little research on \textit{automating the gesture annotation process} that can achieve the two annotation steps simultaneously~\cite{Kratz2015Towards,Ienaga2022Semi}. Meanwhile, the efficient use of rich unlabeled datasets is underexplored.

\begin{figure}
\centering
  \includegraphics[width=0.99\linewidth]{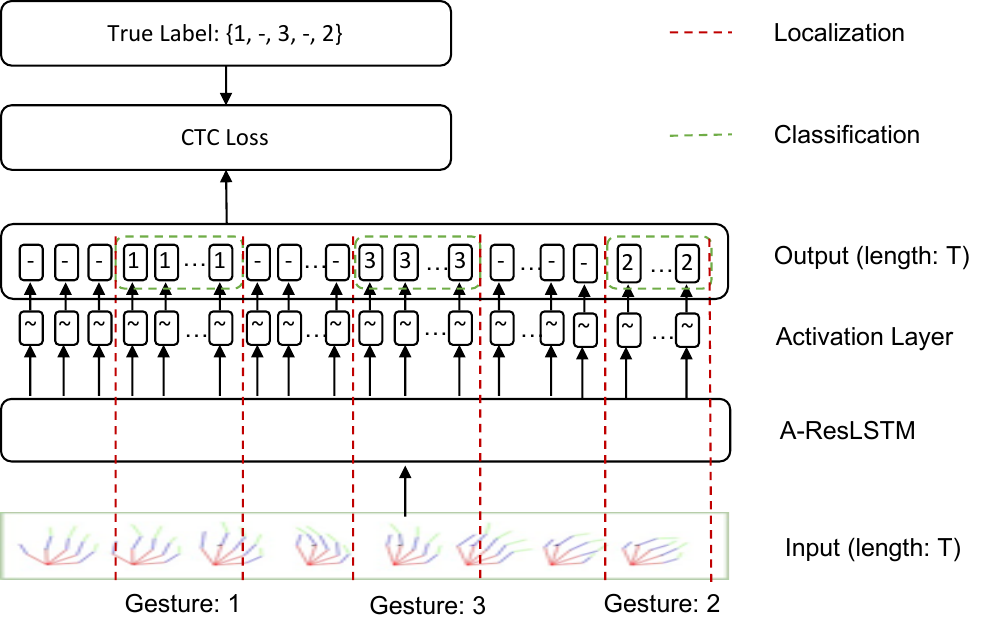}
  \caption{nDue to the extended input window of the annotation model, the true label has the capacity to encompass multiple gestures simultaneously. In the illustrated case, the true labels are \{1, - , 3, - ,2\}, where ``-'' represents background activities (\textit{no gesture}). The output of the annotation model is a sequence of predicted labels that has the same length as inputs. The loss function for the proposed annotation model is a Connectionist Temporal Classification (CTC) loss. The backbone of the model is an A-ResLSTM we adopted from Shen et al.~\cite{Shen2022Gesture}.
  }
  \label{fig:annotation_model}
\end{figure}

To address the gaps, we develop a novel automatic gesture annotation framework (see the red borders in Figure~\ref{fig:illustration}). Our framework contains two components: 1) an annotation model (see Figure \ref{fig:annotation_model}) that enables the prediction of unlabeled gesture data in both gesture classification (step 1 of annotation) and gesture nucleus localization (step 2 of annotation)\footnote{A gesture is composed of three phases: preparation phase, nucleus, and retraction phase. Gesture localization is defined by determining the temporal position of the gesture nucleus, which is the core part of gesture recognition.}.
2) a semi-supervised learning pipeline that uses the annotation model (i.e., the first component) to predict unlabeled data and then uses these predictions as pseudo-labels to further augment the annotation model itself. These high-quality labels can then be used to further fine-tune any downstream gesture recognition models (see the upper part of Figure~\ref{fig:illustration}).

By combining these components, we have developed a robust and automated framework for annotating gesture data of high quality. This framework enables large-scale annotation without the need for extensive human labor. It is important to emphasize that our framework primarily focuses on the annotation step, which serves as a crucial support for training downstream gesture recognition models in real-time systems.

To verify the effectiveness of our framework, we carried out an in-depth analysis using two public gesture-related skeleton-based datasets: SHREC'2021~\cite{caputo2021shrec} and Online DHG~\cite{SHREC2017De}. Initially, we conducted an ablation study to assess the contribution of our different design blocks within the annotation model on gesture classification and nucleus localization tasks. The results showed our model outperforms the baseline model by an improved gesture classification accuracy of 4.3\% (3.4\%) and an improved gesture nucleus localization accuracy of 71.4\% (75.0\%)
in SHREC'2021 dataset (Online DHG dataset).
Next, we used our annotation framework to label a subset of the public datasets mentioned above. We then used this pseudo-labeled dataset, created by our framework, to fine-tune a pre-trained downstream gesture recognition model. The evaluation showed that this pseudo-labeled dataset greatly enhanced the accuracy of the pre-trained downstream gesture recognition model by 11-18\%.

To the best of the author's knowledge, this is the first instance of proposing an automatic and semi-supervised gesture annotation framework that performs gesture classification and nucleus localization simultaneously.

\section{Related Work}
\label{sec:related_work}
We first briefly overview the related work of gesture recognition and localization. We then summarize the existing work on gesture annotation. We also introduce semi-supervised learning background.

\subsection{Gesture Recognition with Deep Learning}

Recently, deep learning techniques have become increasingly popular for gesture recognition. Recurrent neural networks (RNNs) have demonstrated effectiveness in identifying spatial and temporal relationships in activity recognition, both with RGB-based and skeleton-based input~\cite{liu2016spatio, Shen2023FastAR}. However, when there is a large gap between the input data and the target task, conventional RNNs are not capable of extracting relevant information. To address this issue, LSTMs have been proposed to manage `long-term dependencies'\cite{hochreiter1997lstm,Min_2020_CVPR,shahroudy2016ntu}. Additionally, graph and manifold learning has also achieved success in gesture recognition, as seen in DG-STA~\cite{chen2019construct} and ST-TS-HGR-NET~\cite{nguyen2019neural}. There has also been a growing interest in gesture recognition in AR and VR applications, such as 3D hand pose estimation using an infrared camera~\cite{park_3d_2020} and fast recognition of foot gestures for virtual locomotion~\cite{shi_accurate_2019}.

Different modalities of hand gesture data can be used as input for gesture recognition, such as RGB, optical flow, depth, IR-left, IR-disparity, time-series data (IMU, EMG, etc.), non-optical 2D data (tomography, acoustic spectrograms, etc.), and 2D/3D skeletons~\cite{wu_back-hand-pose_2020,yeo_opisthenar_2019,hu_fingertrak_2020,xu_hand_2018,onlinemolchanov2016,mcintosh_sensir_2017,laput_viband_2016,wen_serendipity_2016,escalera_gesture_2017}. 
The framework proposed in this paper can be easily adapted to different input types. As an illustrating example, this paper uses skeleton data, which records the 2D or 3D positions of key points/joints on the hand. With the advancement of hardware (e.g., Microsoft HoloLens, Intel RealSense, and Leap Motion Controller), the use of skeleton data is becoming increasingly popular among different platforms and modalities.
These devices provide precise skeletal data of the hand and fingers in the form of a full 3D skeleton.

\subsection{Gesture Localization}
Previous studies have explored gesture localization using RGB data, with early works such as~\cite{Feng2017511,Mo20061499,Ren2011Robust} primarily focusing on spatial segmentation, which involves separating the hand from its surroundings.
However, it is important to note that temporal segmentation, specifically determining the timing of the gesture nucleus, is equally important in addition to spatial segmentation. This temporal aspect of gesture localization is commonly referred to as ``nucleus localization.''
Existing approaches often employ heuristic methods to locate gestures, such as the use of sliding windows to analyze the sequence of output confidence/loss from a gesture recognition model and identify peaks or valleys~\cite{xu_hulamove_2021, xu_earbuddy_2020}. However, our experiments have shown that such heuristic methods face challenges when dealing with variations in gestures, as depicted in Figure~\ref{fig:loss_function}.

\subsection{Gesture Annotation}
Traditionally, manual annotation software has been employed to annotate gestures, with human annotators manually detecting the start and end points of each gesture. However, this process is time-consuming and requires significant labor.
As a result, there has been growing interest in developing automatic annotation tools. While various fields have explored this idea (e.g.,~\cite{zhang2012review,ramanan2003automatic,duchenne2009automatic}), relatively fewer works have focused on gesture data specifically.
One relevant work by Kratz et al.~\cite{Kratz2015Towards} presents preliminary research on automatically segmenting motion gestures tracked by IMUs. They suggest that by recognizing gesture execution phases from motion data, it might be possible to automatically identify the start and end points of gestures. However, their work primarily focuses on gesture localization and does not address gesture classification.
More closely related to our work, Lenaga et al.\cite{Ienaga2022Semi} propose the use of an Active Learning (AL)~\cite{ren2021survey} framework to automatically detect sign language gesture occurrences in RGB videos. Their approach requires manual annotation for only a small subset of the videos, benefiting researchers studying multimodal communication. However, their primary application area is sign language gestures in RGB videos, which predominantly focuses on spatial data. In contrast, our framework focuses on temporal data for gesture annotation.
Furthermore, our approach leverages a semi-supervised learning pipeline to enhance the performance of annotation, setting it apart from previous works.

\subsection{Semi-Supervised Learning Methods}
In recent years, there has been a growing interest in semi-supervised Learning (SSL) due to its ability to leverage large amounts of unlabeled data. SSL becomes particularly valuable when labeled data is scarce or when the labeling process is time-consuming and labor-intensive. Consistency regularization~\cite{sajjadi2016regularization, bachman2014learning, laine2016temporal} and pseudo-labeling~\cite{lee2013pseudo, xie2020self, rosenberg2005semi} are two popular methods utilized to make effective use of unlabeled data, and they have been integrated into various modern SSL algorithms~\cite{tarvainen2017mean, kurakin2020remixmatch, xie2020unsupervised}.

The semi-supervised learning pipeline within our framework draws inspiration from the recently developed FixMatch algorithm~\cite{sohn2020fixmatch}. FixMatch combines these techniques with both weak and strong data augmentations, yielding remarkable results in SSL tasks.

\section{Problem Formulation}
\label{sec:problem_formulation}
This section formulates gesture annotation and gesture recognition and discusses their difference. We also explain why simply using a conventional gesture recognition model does not suit data annotation and why an annotation model is necessary.

It is important to clarify that when referring to gesture recognition, we specifically focus on \textit{real-time} gesture recognition. In this context, \textit{real-time} gesture recognition involves identifying gestures within a realistic sequence that includes multiple gesture classes as well as background activities (non-gestures), commonly observed in a sensor stream.

To achieve \textit{real-time} gesture recognition, a typical approach is to utilize a sliding-window-based technique. 
At each sliding step, a window is fed into the model, producing class-conditional probabilities. The predicted gesture class within the window is determined by taking the maximum value from the probabilities. However, the conventional real-time gesture recognition model is not well-suited for data annotation, particularly for gesture nucleus localization. Post-processing techniques need to be employed to localize the gesture nucleus. However, these post-processing steps often rely on heuristic methods (e.g.,~\cite{Shen2022Gesture}), which may be susceptible to variations in gestures and contextual differences.

Furthermore, although both gesture annotation and real-time gesture recognition aim for accurate gesture classification, they have different objectives. While real-time gesture recognition models prioritize detecting gestures as quickly as possible, gesture annotation additionally focuses on identifying the nucleus of the gesture as accurately as possible.


In contrast, an annotation model can incorporate gesture nucleus localization as an inherent training objective since it does not need to optimize for recognition latency. This enables better architectural design choices to improve the performance of the annotation task. 

\section{Proposed Framework}

In this section, we describe our framework for annotating gesture data.

\subsection{Design Questions}

We propose an annotation model with individual components that are specifically designed to optimize two goals of gesture annotation: gesture classification and gesture localization. We start by answering four important model design questions of our architecture.


\begin{enumerate}
    \item
\textbf{Small Window Size \textit{vs} Large Window Size:}
The window size for a classification model is typically kept small to ensure accurate classification, as it is believed that a window should only be large enough to contain one gesture~\cite{Shen2022Gesture} such that the input data is clearly segmented and the model can be better trained.
Another main reason for using a small window size is to achieve low recognition latency~\cite{molchanov2016online}, which is not a primary concern during the data annotation process.
Moreover, compared to a larger window, a small window contains less information about previous timestamps and has a narrower vision for gesture localization.
Therefore, we are using a sliding-window technique with a larger window size to process multiple gestures in the input sequence.  

\item 
\textbf{One Prediction Per Frame \textit{vs} Multiple Predictions Per Frame:}
When it comes to identifying gestures in a sliding-window approach, a classification model typically outputs a single vector of class-conditional probabilities for all gesture classes (plus \textit{no gesture}). If the slide step is as small as one frame, the classification model can output a single vector per frame. This is a typical many-to-one mapping, as the input is a sequence of vectors, and the output is one vector.
However, such a many-to-one mapping loses the fine-grained temporal information by collapsing the multiple frames into one single output. This will impact the gesture localization task.
In contrast, a many-to-many (many2many) architecture of each time window contains richer temporal information and allows more dynamic and flexible adjustment between multiple windows.
Therefore, we use a many-to-many model to instantiate such architecture.
The model output is a sequence of class-conditional probabilities and can be decoded as a sequence of predicted classes, with overlapping predictions from consecutive windows.
This allows for dynamic adjustments of earlier predictions based on later windows, and helps to reduce noise in the final prediction.

\item  
\textbf{Strongly-Segmented \textit{vs} Weakly-Segmented Training Data:}
It is important to note that a classification model that is trained using cross-entropy loss typically requires clearly defined and segmented gesture data. This means that the start and end frames of the gesture must be accurately labeled. As a result, the model requires strongly-segmented training data.
However, strongly-segmented data requires extensive human labor. 
We thus desire a model that can be trained using weakly-segmented data, while still being able to learn the temporal and spatial properties of the gesture nucleus among the data.
Therefore, to handle the challenge of the need for strongly-segmented data, we adopt a Connectionist Temporal Classification (CTC) loss, which can automatically align the unsegmented input sequence with the output sequence.
This removes the requirement for strong segmentation of gestural sequences.
This loss has been widely adopted in seq2seq model training~\cite{graves2006connectionist}.

\item
\textbf{Heuristic Parameters \textit{vs} Automation:}
Shen et al.~\cite{Shen2022Gesture} used heuristic threshold parameters to estimate gesture locations.
However, the accuracy of such an estimation largely depends on the other parameters of the model (window length, slide step etc.).
This is because the training label does not directly contain any positional information about the gesture nucleus, and such a conventional real-time gesture recognition model is not explicitly designed for this type of localization.
Therefore, we desire a model that does not need these heuristic parameters. This is another reason supporting our choice of the CTC loss.
The spiky nature of the CTC's output can eliminate the need for post-processing~\cite{vandersteegen2020low,kopuklu2019real}, and allow for the localization of the gesture nucleus without the need for any heuristic parameters. We can identify the nucleus position directly by identifying the spike position. Moreover, greedy search can be used to decode the model output into the sequence of gesture classes.  
\end{enumerate}

\subsection{Annotation Model}
\label{sec:annotation_model}

After answering these four design questions, we now introduce our model's individual components in detail.

\subsubsection{Seq2seq Model}
We design a seq2seq model that is a many2many model. It has an input window $\matr{W}$ of frame length $L$ (window size) and an output which is a matrix of probabilities $\matr{M} \in 
\mathbb{R}^{L\times (K+1)}$.
It defines the probabilities of detecting a gesture (or \textit{no gesture}) $k$ at time $t$ in an input window $\matr{W}$: $p(k,t \mid \matr{W}) = \matr{M}^{t}_{k}\forall t \in \left [ 0,N \right )$.
The backbone of our seq2seq model is an A-ResLSTM model we adopted from Shen et al.~\cite{Shen2022Gesture}, which is one of the state-of-the-art neural network architectures for gesture recognition.
This network is composed of residual blocks~\cite{he2016deep} and bi-directional LSTM layers~\cite{hochreiter1997lstm} with an attention layer~\cite{luong2015effective}. 

\subsubsection{Connectionist temporal classification (CTC)}

\begin{figure}
\centering
\includegraphics[width=0.99\linewidth]{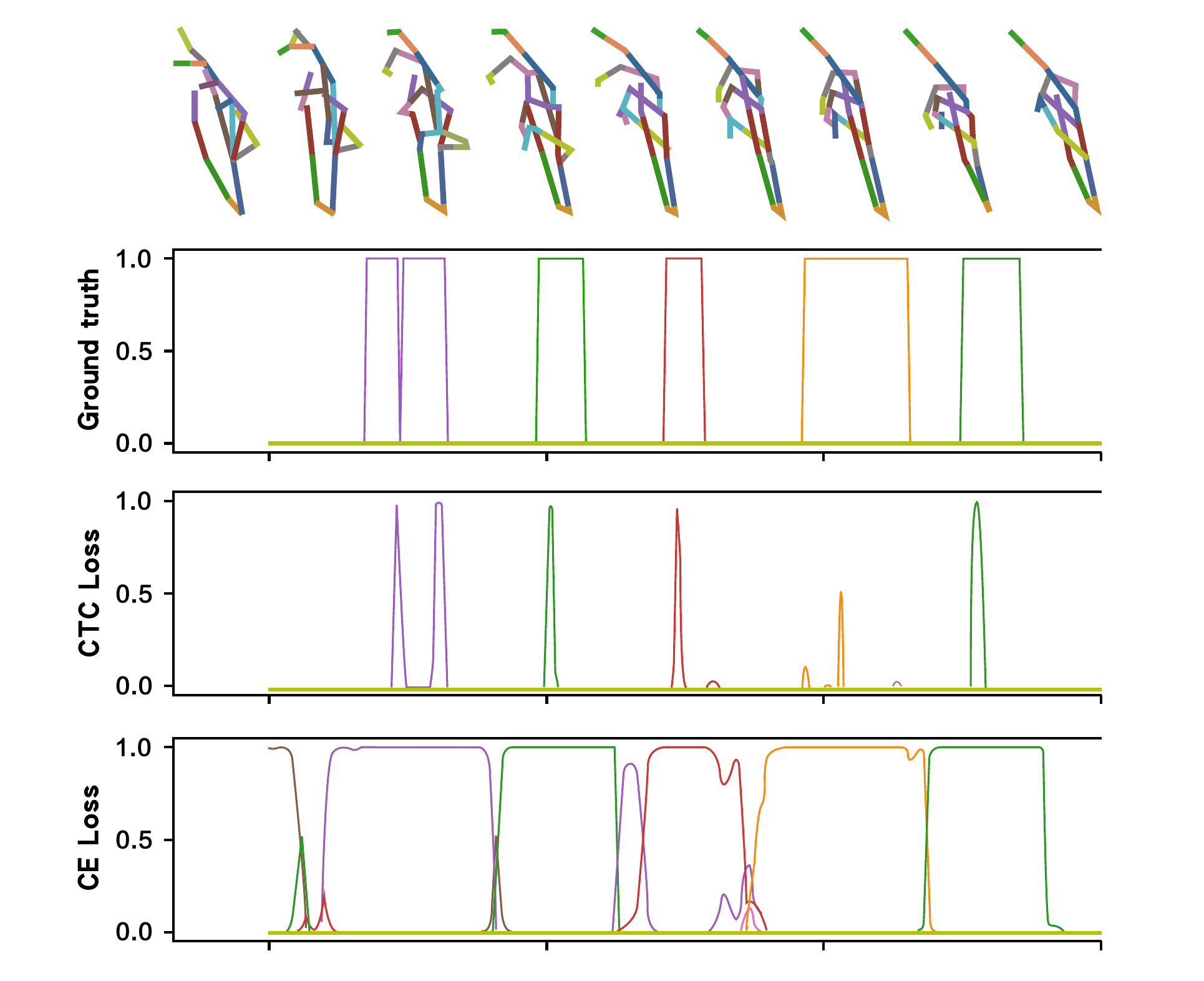}
    \caption{Demonstration of the network output probabilities from a CTC and CE (cross-entropy) trained network versus the ground truth~\cite{vandersteegen2020low}.
    The spike of CTC loss clearly captures the gesture nucleus, while the curve of CE loss needs post-processing.
    For example, the CTC loss successfully distinguishes two consecutive same gestures, while the CE loss confuses them together.}
    \label{fig:loss_function}
\end{figure}

Connectionist temporal classification (CTC) is a type of neural network output that can be constructed into a loss function~\cite{graves2006connectionist}. 
It is designed to tackle the alignment problems between two sequences with very different lengths, such as in handwriting recognition where the written text is often longer than the number of characters. 
This is achieved by introducing a pseudo-character (called blank $-$, or \textit{no label}) to encode duplicate characters in handwriting recognition.
In the gesture recognition case, the pseudo-character is the same as \textit{no gesture}. 
We can then condense repeated class labels and remove \textit{no gesture} labels, e.g., $\left [ -,-,-, 1,1,1,-,2,-,3,3,3,-,- \right ] = \left [ 1, 1,1,-,-,-,2, 2,-,-,3,- \right ] = [1,2,3]$, where 1, 2, 3 are actual gesture classes and ``$-$'' is \textit{no gesture}. 

Mathematically, we define a path $\pi $ as a possible mapping of the input sequence $\matr{W}$ of length $L$ into a sequence of training class labels $\vect{y}$. 
The probability of observing path $\pi$ is $p(\pi \mid \matr{W})  = \prod_{t}^{} \matr{M}_{\pi_{t}}^{t}, \forall \pi \in A'^{L}$, where $\pi_{t}$ is the class label predicted at time $t$ in path $\pi$, and $A'^{L}$ is the set of length $L$ sequences (paths $\pi$) over the gesture dictionary $A' = A\cup \left \{ \textit{no gesture} \right \}$.
The next step is to define a many-to-one mapping operator $y = \gamma(\pi)$ to map the paths into a sequence of gesture labels, $\gamma :  A'^{L} \mapsto A^{\leq L}$, after condensing class labels and removing \textit{no gesture} labels as previously noted.
Thus many paths $\pi$ under $\gamma$ result in the same gesture sequence $\vect{y}$.
The probability of observing the gesture class sequence $\vect{y}$ given an input window $\matr{W}$ is the sum of the condition probabilities of all path $\pi$ mapping to that sequence, $\gamma^{-1}(\vect{y}) = \left \{ \pi: \gamma(\pi) = \vect{y} \right \}$:
\begin{equation}
    p_{CTC}(\vect{y}\mid \matr{W}) = \sum_{\pi \in \gamma^{-1}(\vect{y})}^{}(p(\pi \mid \matr{W}))
    \label{eq:ctc_prob}
\end{equation}
Therefore, the CTC loss is:
\begin{equation}
    L_{CTC} = -log(p_{CTC}(\vect{y}\mid \matr{W}))
    \label{eq:ctc_loss}
\end{equation}
expressed in the log domain. 
$p_{CTC}(\vect{y}\mid \matr{W})$ can be calculated efficiently using a dynamic programming algorithm~\cite{graves2006connectionist}. 
Derivatives of the loss can then be calculated for backpropagation.

Compared to conventional cross-entropy (CE) loss, CTC loss has the benefit of producing output with distinct, sharp predictions that can accurately capture consecutive gestures. In contrast, a model trained with CE tends to blend predictions of the same class together (see Figure~\ref{fig:loss_function}). This means that in order to correctly identify when one gesture has been performed, the output from a CE-trained classification model must be processed further with some heuristic methods with errors (e.g., a single-time activation algorithm~\cite{Shen2022Gesture}). However, with CTC, the sharp nature of the predictions automatically eliminates such error-prone step~\cite{vandersteegen2020low,molchanov2016online, graves2006connectionist}.


\subsubsection{Dynamic Adjustment}
Due to the sliding-window fashion, for a window $\matr{W}$ of length $L$ sliding at step $N$, the model outputs $\matr{M}$ for each window.
$\matr{M}$ consists of $L$ vectors of class-conditional probabilities $\vect{p}$.
Therefore, the model will produce $L/N$ times class-conditional probabilities $\vect{p}$ at each step/frame, forming a sequence $(\vect{p}_{1}, \vect{p}_{2}, ... , \vect{p}_{L/N})$.
We perform a dynamic adjustment on labeling and nucleus localization. 
Specifically, we average these predictions to form one vector of class-conditional probabilities for each step, i.e. $\vect{p}_{average} = \text{average}(\vect{p}_{1}, \vect{p}_{2}, ... , \vect{p}_{L/N})$
We determine the location of the gesture's nucleus by identifying the position of the highest point in the gesture's prediction.

\subsubsection{Decoding}
\label{subsub:decoding}
There are different decoding algorithms for the dynamic adjustment output. 
One option is greedy search, which approximates the solution by taking the most likely class at each step in the matrix $\matr{M}$. It is efficient to compute as it simply concatenates the most active outputs at each step.
Another option is beam search. It uses more information from the output sequence by expanding all possible next steps and keeping track of the $k$ most likely steps, where $k$ is the beam factor, or the maximum number of complexes to be specialized.
It requires more memory and computational power as $k$ increases. But it can generate multiple sequence candidates, which allows for selection based on other contextual information, such as a user's posture.
However, as there is currently no gesture dataset containing such context information, we leave the exploration of beam search or other decoding methods as future work and use greedy search in this paper.

\vspace{0.3cm}
\noindent
Our newly proposed annotation model enables accurate gesture annotations on unlabeled datasets, including both classification and localization. These annotations can be utilized to train various downstream gesture recognition models. However, we also recognize that the potential of unlabeled datasets can be further leveraged. To address this, we introduce a semi-supervised learning pipeline that enhances the performance of both the annotation model and downstream gesture recognition models. This pipeline maximizes the benefits of unlabeled data and improves the overall performance of the framework.

\subsection{Semi-Supervised Learning Pipeline with Pseudo-Labeling}
\label{sec:pipeline}

Once we have an annotation model, we can apply it to unlabeled datasets and use the generated pseudo-labels to further augment the annotation model. This is a technique called pseudo-labeling in semi-supervised learning.
In our case, we design a pseudo-labeling pipeline with a carefully designed training process.
Specifically, we first generate two augmented versions of the same gesture data, one stronger and one weaker. We apply our seed annotation model on the weakly-augmented data and generate pseudo-labels. This is an easier task as the data is less perturbed.
Then, for those labels with high confidence, we assigned them to the strongly-augmented version of the data (a harder task) and further trained the annotation model. By progressing to a more challenging task, the model can further improve its performance.


Formally, let $X = \left \{ (\vect{x}_{b},\vect{y}_{b}:b \in (1,...,B)) \right \}$ be a batch of B labeled examples, where $\vect{x}_{b}$ are the training examples and $\vect{y}_{b}$ are labels. 
Let $U = {\vect{u}_{b}:b\in (1,...,\mu B)} $ be a batch of $\mu B$ unlabeled examples where $\mu$ is a hyperparameter that determines the relative sizes of $X$ and $U$. 
Let $f_{seq2seq}$ denotes our proposed seq2seq model. 
Our approach computes a pseudo-label for each unlabeled sample which is then used in a CTC loss function. 
We perform two types of augmentations: strong and weak, denoted by $\Alpha(\cdot )$ and $\alpha(\cdot)$.

We denote the CTC loss function (not in the negative log domain) as $p_{CTC}(\vect{y}, \vect{x})$ from Equation~\ref{eq:ctc_prob} between label $\vect{y}$ and input $\vect{x}$. 
Let $\matr{M}_{b} = f_{seq2seq}(\alpha(\vect{u}_{b}))$ be the class-conditional probability distribution produced by the model given a weakly-augmented version of a given unlabeled sample: $\alpha(\vect{u}_{b})$.
Then we use $\hat{\vect{y}}_{b} = \textit{path\_search}(\matr{M}_{b})$ as a pseudo-label (\textit{path\_search} is our greedy search algorithm in the decoding step in Sec.~\ref{subsub:decoding}). We enforce the CTC loss against the model's output for a strongly-augmented version of the unlabeled sample:
$\Alpha(\vect{u}_{b})$:
\[
    L = \frac{1}{\mu B}\sum_{b= 1}^{\mu B}\mathbb{I}(\textit{LP\_path\_search}(q_{b})\geq \tau )p_{CTC}(\hat{\vect{y}}_{b}\mid , \Alpha(\vect{u}_{b}))
\]
where $\textit{LP\_path\_search}$ is the log-probabilities of the best path in greedy search, and $\tau$ is the threshold, 

We employ basic time-series transformations~\cite{iwana_empirical_2021} for data augmentation, which include scaling, shifting, time interpolation, and adding noise. These transformations help to diversify the training data and enhance the robustness of the model. For weak augmentation, we only apply noise; for strong augmentation, we apply all the transformations.

The threshold for determining if the pseudo-labels should be remained and be used to update the model controls the trade-off between the quality and the quantity of pseudo-labels.
Recent work suggests that the quality of pseudo-labels is more important than the quantity for a good performance~\cite{sohn2020fixmatch}. Thus we empirically set the threshold to 0.9.

\section{Experiments and Analysis}
\renewcommand{\arraystretch}{1.3}
\begin{table}[t]
\centering
\resizebox{\columnwidth}{!}{
\begin{tabular}{lll}
\hline\hline
                 & \makecell{\textbf{SHREC'2021}~\cite{caputo2021shrec}}                           & \makecell{\textbf{Online DHG}~\cite{SHREC2017De}}                                                                                                                \\ \hline
Static           & one - four, OK, menu      & NA                                                                                                                          \\ \hline
Dynamic          & left, right, circle, v, cross        & \makecell[l]{rotation (counter-)clockwise,\\ right/left,  up/down,\\ cross, plus, v, shake} \\ \hline
Fine Dynamic     & \makecell[l]{grab, pinch, tab, deny,\\ knob, expand} & \makecell[l]{grab, pinch, tab, expand}                                                                                                    \\ \hline
Size     & \makecell[l]{180 seqs of 3-5 gestures\\ occurring sequentially} & \makecell[l]{280 seqs of 10 gestures\\ occurring sequentially}                                                                             \\ \hline\hline
\end{tabular}
}
\caption{Contents of the datasets. Static gestures are characterized by keeping a fixed hand pose for a minimum amount of time, while dynamic gestures are characterized by a single trajectory with an unchanged hand pose or with finger articulation over time.
Fine dynamic gestures are characterized by a single trajectory with changing hand pose. }
\label{tab:dataset}
\end{table}
\renewcommand{\arraystretch}{1.0}

\renewcommand{\arraystretch}{1.3}
\begin{table*}[t]
\centering
\begin{tabular}{lcccc}
\hline \hline
      & \multicolumn{2}{c}{\textbf{SHREC’2021~\cite{caputo2021shrec}}} & \multicolumn{2}{c}{\textbf{Online DHG~\cite{SHREC2017De}}} \\ \hline
      & Accuracy           & NNLE           & Accuracy           & NNLE           \\ \hline
A-ResLSTM~\cite{Shen2022Gesture} w/ CE Loss (Baseline Model) &      88.3 ($\pm$ 2.34)         &       0.42 ($\pm$ 0.07)        &     89.8 ($\pm$ 3.12)          &          0.44 ($\pm$ 0.03)        \\ \hline
A-ResLSTM w/ CTC loss  &         90.6 ($\pm$ 1.45)         &       0.17 ($\pm$ 0.03)        &     91.5 ($\pm$ 3.35)          &          0.16 ($\pm$ 0.04)            \\
A-ResLSTM w/ CTC loss \& many2many  &     91.9 ($\pm$ 1.12)         &       0.14 ($\pm$ 0.03)        &     92.7 ($\pm$ 3.35)          &          0.15 ($\pm$ 0.04)            \\ \hline
A-ResLSTM w/ CTC loss \& many2many \& dynamic adjustment  &       \textbf{92.6 ($\pm$ 1.58)}        &   \textbf{0.12 ($\pm$ 0.02)}           &      \textbf{93.2 ($\pm$ 1.76)}         &          \textbf{0.11 ($\pm$ 0.03)}        \\ \hline \hline
\end{tabular}
\caption{
Ablation study results by adding our multiple design parts step by step.
NNLE is our metric for gesture localization (\textbf{N}ormalized \textbf{N}ucleus \textbf{L}abeling \textbf{E}rror).
The numbers in the brackets indicate std.}
\label{tab:baseline_comparisons}
\end{table*}
\renewcommand{\arraystretch}{1.0}

We conducted a series of experiments to evaluate our proposed annotation model and the framework.

\subsection{Evaluation Datasets}
In order to ensure a thorough evaluation of our framework, we perform our experiments on two publicly available datasets. These datasets consist of sequences of unsegmented gesture data, encompassing various types of gestures. The specific details of the two datasets are as follows:

\begin{enumerate}
    \item \textbf{SHREC'2021~\cite{caputo2021shrec}}:
    The dataset utilized in this study consists of 18 distinct gesture classes belonging to various types. It comprises a total of 180 gesture sequences, each carefully designed to incorporate 3 to 5 gestures, accompanied by additional semi-random hand movements labeled as non-gestures. The original dictionary comprises 18 gestures, encompassing both static and dynamic gestures.
    \item \textbf{Online DHG~\cite{SHREC2017De}}:
    The dataset used in this study comprises 14 distinct gesture classes from various categories. It consists of 280 sequences where each sequence contains 10 unsegmented gestures occurring sequentially. The data in this dataset is represented in the form of skeleton data, specifically capturing 22 joints for each hand skeleton. The skeleton data was collected using a Leap Motion device. 
\end{enumerate}

Both datasets already provided a predefined split between training and testing data. Thus, we conducted evaluations on the designated testing sets.

\subsection{Evaluation Metrics}
We introduce our metrics to evaluate the tasks of gesture classification and localization.

\begin{enumerate}
\item \textbf{Gesture Classification Accuracy}:
Gesture annotation would generate a sequence of classification output.
We use the Levenshtein distance (also known as minimum edit distance) to evaluate the gesture classification performance~\cite{kopuklu2019real}.
It is a metric defined as the minimum number of single-character (in our case, classification output of each frame) insertions, deletions, and substitutions required to transform one string into another. 
The accuracy for recognition performance is defined as:
\begin{equation}
    1-\frac{\operatorname{levenshtein}(y_{predict},y_{true})}{\operatorname{length}(y_{true})}
    \label{eq:accuracy}
\end{equation}
where $y_{predict}$ and $y_{true}$ are the predicted and true list of labels of the gestures, respectively.

\item \textbf{Nucleus Localization Error}:
We propose a metric Normalized Nucleus Localization Error (NNLE) to measure how accurately our model can locate the gesture nucleus.
When a gesture is recognized, we define the time for the start/end of the gesture as $idx_{start}$/$idx_{end}$, and the detected location of gesture nucleus as $idx_{nucleus}$. 
For a accurate nucleus localization, $idx_{start} \leq idx_{nucleus} \leq idx_{end}$. 
NNLE is defined as:
\begin{equation}
    \frac{idx_{activation}-(idx_{start}+idx_{end})/2+1}{idx_{end}-idx_{start}+1}
    \label{eq:NNLE}
\end{equation}
A smaller NNLE means our annotation model is more accurate for gesture localization (i.e., the nucleus is closer to the center of the gesture).
\end{enumerate}

\subsection{Training Details}
Our implementation was based on TensorFlow 2. We utilized the Adam optimizer~\cite{Kingma2014AdamAM} with a starting learning rate of 0.0001.
To prevent overfitting, we employed early stopping with a patience of 5.
For both the Online DHG and SHREC'2021 datasets, we used a window length of 200 for our model. It is important to note that the length of an individual gesture within the datasets ranged from 20 to 50 frames, corresponding to a duration of approximately 0.4 to 1 second.

\subsection{Ablation Study}
\label{sec:annotation_model_evaluation}


Our novel annotation model essentially consists of multiple design blocks: (1) using the CTC loss instead of a basic CE loss, (2) using a many-to-many architecture (i.e., seq2seq) instead of a many-to-one architecture, and (3) using dynamic adjustment on labeling and nucleus localization. To evaluate the effectiveness of each design block, we performed an ablation study through a step-by-step addition process.

Table~\ref{tab:baseline_comparisons} illustrates the ablation study results.
Compared to the baseline (A-ResLSTM trained with Cross-Entropy (CE) loss), our model significantly reduces the gesture localization error by 0.30 (71.4\%) and 0.33 (75.0\%).
Regarding gesture classification accuracy, our model design achieves improvements of 4.3\% on the SHREC'2021 dataset and 3.4\% on the Online DHG dataset compared to the baseline.

Each of the three blocks contributes approximately equally to the overall gesture classification accuracy improvement. 
For gesture nucleus localization, the main contribution stems from the application of the Connectionist Temporal Classification (CTC) loss. 
This highlights the unique character of the CTC-trained model in producing outputs with a spiky nature, eliminating the need for complex post-processing techniques (see Figure~\ref{fig:loss_function}). In contrast, CE-trained networks require post-processing to determine the final label, often relying on heuristic methods.

\subsection{Annotation Framework Evaluation}

\renewcommand{\arraystretch}{1.3}
\begin{table*}[t]
\centering
\begin{tabular}{lcccccc}
\hline \hline
               & \multicolumn{3}{c}{\textbf{SHREC'2021~\cite{caputo2021shrec}}}                                     & \multicolumn{3}{c}{\textbf{Online DHG~\cite{SHREC2017De}}}                                     \\ \hline
Labeled Sequences         & $\sim40$                 & $\sim80$                 & $\sim120$                 & $\sim70$                & $\sim140$                & $\sim210$                 \\ \hline
Baseline~\cite{Shen2022Gesture}           & \textbf{30.0 ($\Delta$=22\%)}          & 69.0 ($\Delta$=11\%)          & 83.3 ($\Delta$=6\%)           & \textbf{34.9 ($\Delta$=25\%)}          & 68.3 ($\Delta$=10\%)          & 81.6 ($\Delta$=5\%)           \\ 
Our Annotation Framework & 26.0 ($\Delta$=18\%) & \textbf{73.0 ($\Delta$=15\%)} & \textbf{88.3 ($\Delta$=11\%)} & 28.9 ($\Delta$=19\%) & \textbf{75.3 ($\Delta$=17\%)} & \textbf{89.6 ($\Delta$=13\%)} \\ \hline \hline
\end{tabular}
\caption{Evaluation of the effectiveness of the pseudo-labeled dataset produced from our proposed annotation framework.
The numbers indicate the gesture classification accuracy of the downstream gesture recognition model fine-tuned from the pseudo-labeled dataset. The $\Delta$ indicates the accuracy improvement between the pre-trained accuracy and the fine-tuned accuracy. The second row indicates the size of the labeled subset. These labeled sequences are used for pre-training the downstream gesture recognition model and the annotation model. Please refer to Figure~\ref{fig:illustration} for a detailed illustration of the workflow of the annotation framework. Note that SHREC'2021 dataset has a total of 180 sequences, and Online DHG dataset has a total of 280 sequences.
}
\label{tab:model_self_training}
\end{table*}
\renewcommand{\arraystretch}{1.0}

To evaluate the effectiveness of our framework, we incorporate a similar pseudo-labeling process into the baseline model described in Section~\ref{sec:annotation_model_evaluation}. This modified version serves as a baseline framework for comparison.

\subsubsection{Baseline}

It is important to note that the pseudo labeling semi-supervised learning pipeline we developed specifically for our proposed annotation model cannot be directly applied to the baseline model, as the baseline model is trained using Cross-Entropy (CE) loss. Therefore, we made modifications to adapt the pseudo-labeling process to the baseline model, enabling a fair comparison between the two frameworks.

Formally, we denote the CE loss function between two probability distributions $p$ and $q$ as $H(p,q)$. 
Let $q_{b} = p(\vect{y} \mid \alpha(\vect{u}_{b}))$ be the class-conditional probability distribution produced by the model given a weakly-augmented version of a given unlabeled sample: $\vect{u}$.
Then we use $\hat{q}_{b} = \argmax(q_{b})$ through the path decoding algorithm as a pseudo-label, and we enforce the CE loss against the model's output for a strongly-augmented version of $\vect{u}_{b}$:
\[
    L = \frac{1}{\mu B}\sum_{b= 1}^{\mu B}\mathbb{I}(\argmax(q_{b})\geq \tau )H(\hat{q}_{b}\mid ,p(\vect{y}\mid \Alpha(\vect{u}_{b})))
\]

where $\tau$ is the threshold. The rest of the semi-supervised learning pipeline stays the same as in Section~\ref{sec:pipeline}.

\subsubsection{Procedure}
The ultimate objective of our framework is to generate high-quality pseudo-labels that can effectively fine-tune and enhance downstream gesture recognition models. To evaluate the efficacy of our pipeline, we measure the performance improvement achieved by fine-tuning the initial pre-trained gesture recognition model using our generated pseudo-labels.

In the experiments, we create a subset of the training dataset that initially contains complete labels. From this subset, we remove the labels and leverage our framework to generate pseudo-labels for this particular portion of the data.
The size of the remaining subset of labeled data becomes the adjustable parameter in our experiments. 
Now, we have two subsets, one labeled subset and one unlabeled subset. 
This labeled subset is utilized for pre-training both the annotation model and the downstream gesture recognition model, and the unlabeled subset is utilized in the pseudo labeling-based semi-supervised learning pipeline. Such a process is demonstrated in Figure~\ref{fig:illustration}. We use one of the state-of-the-art gesture recognition models, ST-GCN, from~\cite{caputo2021shrec} as the downstream gesture recognition model. 
The performance of the framework directly impacts the quality of the pseudo-labels produced from the annotation framework.
The quality of the pseudo-labels then impacts the improved accuracy and final accuracy of the final downstream recognition model which is fine-tuned from the pseudo-labels. This is explained in Figure~\ref{fig:illustration}. We use gesture classification accuracy in Equation~\ref{eq:accuracy} to measure the accuracy of the downstream gesture recognition model.


\subsubsection{Results}
The performance and improvements of the fine-tuned downstream recognition model are summarized in Table~\ref{tab:model_self_training}. It is important to note that these results pertain to the downstream model and are not directly comparable to those in Table~\ref{tab:baseline_comparisons}.

Through the semi-supervised learning pipeline, both our new annotation framework and the baseline framework demonstrate significant improvements in the performance of the downstream model, as indicated by the $\Delta$ values, from 5\% to 25\%. The magnitude of improvement is directly influenced by the size of the unlabeled dataset, with larger datasets yielding greater improvements through this pipeline.

Furthermore, our evaluation results highlight that the pseudo-labeled data generated by our framework generally outperforms the pseudo-labeled data generated by the baseline framework. The quality of the pseudo labels is reflected in both the improved accuracy and the final accuracy of the downstream model.
It is noteworthy that the baseline framework initially outperforms our new framework when using a small set of labeled data. This can be attributed to the higher number of trainable parameters in the output layer of our annotation model, which employs a many2many design and faces challenges in achieving convergence with the CTC loss. However, as the size of the labeled set increases, our new framework surpasses the baseline.

The superior performance of our framework can be attributed to the annotation model's ability to train on pseudo-labeled data that may not be accurately segmented initially in the semi-supervised learning loop (pipeline). This is in contrast to the baseline framework, whose model heavily relies on accurately segmented data when trained with CE loss.
These results highlight the potential of our annotation model and the semi-supervised pipeline to significantly enhance downstream gesture recognition models, showcasing the advantages of our approach.

\section{Limitation and Future Work}

It is important to acknowledge several limitations in our work. Firstly, although our proposed model exhibits superior performance in gesture data annotation for class labeling and nucleus localization, we must acknowledge that the training process of our new model is more time-consuming compared to the baseline. While training efficiency is not the primary focus since annotation can occur offline, there is room for future improvement in this aspect.
Furthermore, our current framework still relies on the availability of labeled data to initiate the process. We have not explored a fully unsupervised version in this study, which presents an intriguing avenue for future research.

\section{Conclusion}
This paper presents a framework that enables accurate and automated annotation of gesture data, encompassing gesture classification and localization. The framework consists of two key components: (1) a novel annotation model that optimizes both gesture classification and localization concurrently, and (2) a semi-supervised learning pipeline incorporating pseudo-labeling.
The ablation study reveals that this novel annotation model design surpasses the baseline model, achieving a 4.3\% higher class labeling accuracy and a 71.4\% improvement in nucleus localization accuracy on the SHREC'2021 dataset (3.4\% and 75.0\% respectively on the Online DHG dataset). Moreover, we also demonstrate that the pseudo-labeled data generated by our framework significantly enhances the performance of a pre-trained downstream gesture recognition model through fine-tuning, resulting in improvements ranging from 11\% to 18\% on the SHREC'2021 and Online DHG datasets.




{\small
\bibliographystyle{ieee}
\bibliography{egbib}
}

\end{document}